\documentclass[11pt]{article}
\usepackage{coling2020}
\usepackage{times}
\usepackage{url}
\usepackage{latexsym}
\usepackage{graphicx}

\colingfinalcopy 

\title{Separating Argument Structure from Logical Structure in AMR}

\author{Johan Bos\\
        University of Groningen\\ 
        \texttt{johan.bos@rug.nl}}

\date{}

\begin{document}
\maketitle

\newcommand{\omr}[2]{\tt (#1 / #2)}
\newcommand{\amr}[4][]{\tt (#2 /#3/ #4)#1}
\newcommand{\namr}[3]{\tt (#1 $\stackrel{\hspace*{3pt}[#2]}{/}$ #3)}
\newcommand{\rol}[3]{\\ \hspace*{#1pt}:#2 #3}
\newcommand{\nrol}[3]{\\[-7pt] \hspace*{#1pt}:#2 #3}
\newcommand{\con}[1]{\begin{tabular}{r}#1\\~\\\end{tabular}}

\newcommand{\drs}[2]
{
    \begin{tabular}{|c|}
    \hline
    #1
    \\
    ~ \vspace{-3ex} \\
    \hline
    ~ \vspace{-2ex} \\
    #2
    \\
    \hline
    \end{tabular}
}

\newcommand{\flatdrs}[2]
{
    \begin{tabular}{|c|c|}
    \hline
    #1 & #2 \\
    \hline
    \end{tabular}
}

\begin{abstract}
The AMR (Abstract Meaning Representation) formalism for representing
meaning of natural language sentences puts emphasis on predicate-argument structure and was not designed to deal with
scope and quantifiers.  By extending AMR with indices for contexts and
formulating constraints on these contexts, a formalism is derived that
makes correct predictions for inferences involving negation and bound
variables.  The attractive core predicate-argument structure of AMR is
preserved. The resulting framework is similar to the meaning representations of Discourse
Representation Theory employed in the Parallel Meaning Bank.
\end{abstract}

\section{Introduction}

Abstract Meaning Representation, AMR \cite{LangkildeKnight}, or the PENMAN notation it is based on \cite{Kasper:1989}, puts emphasis on argument
structure. In this paper I put forward a proposal to extend AMRs with a
logical dimension, in order to---from a formal semantics point of
view---correctly capture negation, quantification, and
presuppositional phenomena.  
It is desirable to investigate such an extension, because 
(i) it would make a comparison of AMR with other semantics formalisms possible (in particular Discourse Representation Theory);
(ii) it would make AMR suitable for performing logical inferences; and 
(iii) it would be an important step in sharing resources for semantic parsing. 
The aim is to do this in such a way that existing AMR-annotated corpora \cite{amr} can be
relatively easily extended with the desired extensions.

This is not the first proposal of extending the PENMAN notation to handle scope
phenomena. The need to do so was recognized by other researchers 
 \cite{Bos2016CL,Stabler:2017,Lai2020}. \newcite{pustejovsky-etal-2019-modeling} extend AMR with a possibility of
adding explicit scope relations. This extension, however, doesn't
solve a fundamental problem that AMR faces, namely viewing AMRs as directed acyclic graphs and basing 
 their interpretation on this.
Consider examples such as
``every snake bit itself'' or ``all dogs want to swim''. In the original AMR graph notation,
where quantifiers are expressed as a predicate rather than taking
scope, the resulting diagrams are:

%
\begin{quote}
  \hfill
\includegraphics[height=55mm]{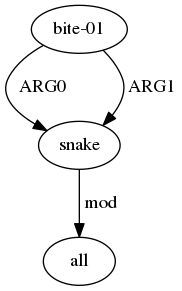}\hfill
\includegraphics[height=60mm]{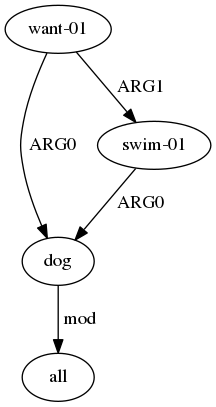}\hfill ~
\end{quote}

The corresponding interpretations for these sentences could be
paraphrased as ``every snake bit every snake'' and ``all dogs want all
dogs to swim'', which are not the meanings that the sentences
express. Let's refer to this issue as the \textit{bound variable problem}.


But there is a second, perhaps an even more pressing issue, which could be
dubbed the \textit{scope representation problem}.  AMRs, in their
original PENMAN format \cite{Kasper:1989}, cannot be used directly for
drawing valid inferences. Let me demonstrate why this is so. Using the
simple conjunction elimination rule (if the conjunction "A and B" is
true, then "A" is true, and "B" is true), and assuming that an AMR is
interpreted as a conjunction of clauses, AMR will make the right
predictions as long as no negation is involved (e.g., it will yield
the correct inference ``Mary left'' from ``Mary left yesterday''). But
since, in AMR, negation is represented as a predicate rather than an
operator that takes scope, it will make wrong predications for negated
sentences (e.g., it will allow the inference ``Mary left'' from ``Mary
did not leave''. This is why AMRs need some kind of reformulation
before interpretation, and that is exactly what has been proposed in earlier work \cite{artzi-etal-2015-broad,Bos2016CL,Stabler:2017,Lai2020}. 

In this paper I seek the solution at the
representational level. I think the contribution of this paper is that this
extension is simpler of nature than those proposed earlier \cite{Bos2016CL,Stabler:2017}. It bears similarities with \textit{named graphs}
for semantic representations \cite{Crouch:2018}. I argue that, if we want to fix the bound variable problem and the
scope representation problem, AMR requires explicit scope in their
representations \cite{30musts}.  I propose a
method to do this by keeping the underlying predicate-argument
structure, and adding a second, logical layer
(Section~\ref{sec:method}).  In Section~\ref{sec:results}, a
list of examples of extended AMRs demonstrate the approach. Some
loose ends are discussed in Section~\ref{sec:discussion}.

\bigskip

\section{Method}\label{sec:method}

The idea is to extend the AMR with logical structure, obtaining a
scoped representation AMR$^+$ with two dimensions: one level
comprising predicate-argument structure (the original AMR, minus
polarity attributes), and one level consisting of the logical
structure (information about logical operators such as negation and
the scope they take). This is achieved by viewing an AMR as a recursive
structure, rather than interpreting it as a graph, and performing two
operations on them:

\begin{enumerate}

\item assign an index to each (sub-)AMR;

\item add structural constraints to the AMR via the indices.
  
\end{enumerate}

AMRs can be seen as a recursive structure by viewing every slash
within an AMR as a sub-AMR \cite{Bos2016CL}.  If a (sub-)AMR contains
relations, those relations will introduce nested AMRs. (A constant is
also an AMR, following this view.)  An AMR (and all its sub-AMRs) will
be labeled by decorating the slashes with indices (indices will be
indicated by numbers enclosed in square brackets).

Every AMR is augmented by a set of scoping constraints on the labels.
This way, a sub-AMR can be viewed as describing a ``context''. The
constraints state how the contexts relate to each other. They can be declared as
the same contexts (=), a negated context ($\lnot$), a conditional
context ($\Rightarrow$), or a presuppositional context ($<$).  
Colons
are used to denote inclusion, i.e., $l:C$ states that context $l$ contains
condition $C$.  

Note that these labels are similar in spirit to those used in
underspecification formalisms as proposed in the early 1990s \cite{reyle:udrs,CopestakeEA:mrs,bos:plu}. 
 The treatment of presuppositions is inspired by semantic formalism extending Discourse Representation Theory
\cite{sandt:presupdrt,geurts:1999,pdrt,Venhuizen2018JofS}.

\bigskip

\section{Results}\label{sec:results}

Below I illustrate the idea with several canonical examples involving
existential and universal quantification, definite descriptions,
proper names, and, of course, negation. 

\subsection{Existential Quantification}

Consider the AMR for ``a dog scared a cat'' with a transitive verb
and two indefinite noun phrases in PENMAN notation:

\begin{quote}
\omr{e}{scare-01 \rol{19}{ARG0}{\omr{x}{dog}} 
                 \rol{19}{ARG1}{\omr{y}{cat}}}    
\end{quote}

Within this AMR
 we can identify three
sub-AMRs. We index and constrain them and arrive at the following AMR$^+$:

\begin{quote}
\amr[ $\{1=2,1=3\}$]{e}{1}{scare-01 \rol{26}{ARG0}{\amr{x}{2}{dog}} 
                          \rol{26}{ARG1}{\amr{y}{3}{cat}}}    
\end{quote}

Here, there is just one context shared by all three sub-AMRs, as one
would expect with existential quantification: scope does not play a pivotal
role here. As equivalent alternative, the following simplified,
constraint-free AMR$^+$ can be obtained after eliminating the identity
constraints:

\begin{quote}
\amr[ $\{\}$]{e}{1}{scare-01 \rol{26}{ARG0}{\amr{x}{1}{dog}} 
                          \rol{26}{ARG1}{\amr{y}{1}{cat}}}    
\end{quote}

\subsection{Definite Descriptions and Proper Names}

A sentence like ``the bear growled'' contains a definite description
triggering an existential presupposition. Presuppositions yield new contexts:

\begin{quote}
\amr[ $\{$2$<$1$\}$]{e}{1}{growl-01 \rol{26}{ARG0}{\amr{x}{2}{bear}}} 
\end{quote}

In other words, the definite article triggers a presupposition that
there is a bear (the AMR with index 2) with respect to context
provided by the AMR indexed as 1. Proper names can be handled
similarly, as the AMR$^+$ for the sentence ``Fido barked'' shows (the existence of a dog named ``Fido'' is a presupposition for the barking event):

\begin{quote}
\amr[ $\{$2$<$1$\}$]{e}{1}{bark-01 \rol{26}{ARG0}{\amr{x}{2}{dog \rol{90}{Name}{"Fido"}}}}
\end{quote}

\subsection{Negation}\label{sec:negation}

Negation introduces a new (negated) context in AMR$^+$. This makes the
\texttt{:polarity-} relation in AMR obsolete.  As negation is always
part of another context in some cases a context needs to be coerced
(second and third example below, see also Section~\ref{sec:inferred}).  Consider
the representations for "a woman didn't smile", "the woman didn't
smile", and "no woman smiled":

\begin{quote}
\amr[ $\{$2:$\lnot$1$\}$]{e}{1}{smile-01 \rol{26}{ARG0}{\amr{x}{2}{woman}}}
\end{quote}

\begin{quote}
\amr[ $\{$3$<$1,2:$\lnot$1$\}$]{e}{1}{smile-01 \rol{26}{ARG0}{\amr{x}{3}{woman}}}
\end{quote}

\begin{quote}
\amr[ $\{$2:$\lnot$1$\}$]{e}{1}{smile.v.01 \rol{26}{Agent}{\amr{x}{1}{woman}}}
\end{quote}

\subsection{Universal Quantification}\label{sec:quant}

Universal quantification introduces a conditional context in AMR$^+$. Below are examples for quantifiers in subject position, object position, subject and object position, and a quantification with a bound variable. Note that the \textit{mod}-relation used in AMR becomes obsolete.

\begin{quote}
``Everyone smiled.''\\
\amr[ $\{$3:2$=>$1$\}$]{e}{1}{smile-01 \rol{26}{ARG0}{\amr{x}{2}{person.n.01}}}
\end{quote}

\begin{quote}
``A dog scared every cat.''\\
\amr[ $\{$3:2$=>$1$\}$]{e}{1}{scare-01 \rol{26}{ARG0}{\amr{x}{3}{dog}} \rol{26}{ARG1}{\amr{y}{2}{cat}}}  
\end{quote}

\begin{quote}
``Every dog scared every cat.''\\
\amr[ $\{$5:3$=>$4,4:2$=>$1$\}$]{e}{1}{scare-01 \rol{26}{ARG0}{\amr{x}{2}{dog.n.01}} \rol{26}{ARG1}{\amr{y}{3}{cat}}}
\end{quote}

\begin{quote}
``Every student revised their paper.''\\
\amr[ $\{$2$=$3,3$<$1,4:2$=>$1$\}$]{e}{1}{revise-01 \rol{26}{ARG0}{\amr{x}{2}{student}} 
                                            \rol{26}{ARG1}{\amr{y}{3}{paper \rol{98}{poss}{x}}}}
\end{quote}

\subsection{From AMR to DRS}\label{sec:translation}

The AMR$^+$ representations share characteristics with the Discourse
Representation Structure (DRS) introduced in Discourse Representation
Theory, DRT for short \cite{kampreyle:drt}.  It is important to
compare AMR$^+$ with DRS for various reasons. DRT is a well-studied
formalism with a model-theoretic component. If we are able to show
that the representations are equivalent then this has positive
consequences for AMR, as all inferential properties supplied by DRT
could be transferred to AMR.

As a matter of fact, there is a rather straightforward way of
converting labelled AMRs to DRS in the style of the Parallel Meaning
Bank \cite{pmb}. This conversion comprises three main steps
($\tau$ is the  translation function, $\oplus$ is a
DRS-merge operation, and v is a function mapping an AMR to its main variable):

\begin{enumerate}

\item Replace each sub-AMR by a DRS. This DRS contains exactly one
  discourse referent, a one-place predicate, and zero or more
  two-place relations. The first argument of the two-place relation is
  the main variable of the sub-AMR; the second argument of the
  two-place relation is the main variable of the sub-AMR. So given an
  AMR$^+$ (x/i/C :R$_1$ A$_1$ ... R$_n$ A$_n$), the corresponding DRS is
    $\tau$(i) = \flatdrs{x}{C'(x) R'$_1$(x,v(A$_1$)) ... R'$_n$(x,v(A$_n$))}.

\item Merge all DRSs that are indexed with the same index. A merge of
  two DRSs ($\oplus$) consists of taking the unions of their respective domains
  and conditions.  Assign an empty DRS \flatdrs{}{} to inferred
  contexts.
  
\item Construct the final DRS by following the structure expressed by
  the constraints.
  For instance, 3:2$=>$1 is translated as $\tau$(3) $\oplus$ \flatdrs{}{$\tau$(2) $\Rightarrow$ $\tau$(1)},
  and 1:$\lnot$2 is translated as $\tau$(1) $\oplus$ \flatdrs{}{$\lnot$ $\tau$(2)}.

\end{enumerate}

Here are two examples
that illustrate this translation, converting AMR
predicates to PMB WordNet synsets, and AMR PropBank relations to PMB
VerbNet roles:

\begin{quote}
"A dog scared every cat."\\
\amr[ $\{$3:2$=>$1$\}$]{e}{1}{scare-01 \rol{26}{ARG0}{\amr{x}{1}{dog}} \rol{26}{ARG1}{\amr{y}{2}{cat}}}  
\end{quote}

\begin{quote}
$\tau$(1) = \drs{e}{scare.v.01(e)\\Stimulus(e,x)\\Experiencer(e,y)} $\oplus$ \drs{x}{dog.n.01(x)} = \drs{e x}{scare.v.01(e)\\Stimulus(e,x)\\Experiencer(e,y)\\dog.n.01(x)}

$\tau$(2) =  \drs{y}{cat.n.01(y)}

$\tau$(3) = \drs{}{$\tau$(2) $\Rightarrow$ $\tau$(1)} = \drs{}{\drs{y}{cat.n.01(y)}$\Rightarrow$\drs{e x}{scare.v.01(e)\\Stimulus(e,x)\\Experiencer(e,y)\\dog.n.01(x)}\\[-9pt]}
\end{quote}

\begin{quote}
"Mary didn't smile."\\
\amr[ $\{$2$<$3,3:$\lnot$1$\}$]{e}{1}{smile-01 \rol{26}{ARG0}{\amr{x}{2}{person.n.01 \rol{100}{Name}{"Mary"}}}}
\end{quote}

\begin{quote}
$\tau$(1) =  \drs{e}{smile.v.01(e)\\Agent(e,x)}

$\tau$(2) = \drs{x}{person.n.01(x)\\Name(x,"Mary")}  

$\tau$(3) =  \drs{}{$\lnot$$\tau$(1)} = \drs{}{$\lnot$ \drs{e}{smile.v.01(e)\\Agent(e,x)}\\[-9pt]}
\end{quote}

In terms of expressive power, AMR$^+$ is equivalent with the dialect
of DRS employed in the Parallel Meaning Bank \cite{pmb}, where
relations cannot have arity larger than two.  In general, DRS is a
more expressive meaning representation language.

\section{Discussion}\label{sec:discussion}

In this section I discuss some loose ends that emerged when designing AMR$^+$: the issue of inferred
labels, annotation work required to implement the approach, 
and the conversion to triples.

\subsection{Inferred Labels}\label{sec:inferred}

In the current proposal, negation and conditionals introduce new
indices, that do not appear in the predicate-argument level of the
AMR. These are necessary to ensure a well-formed logical
structure. But they are (perhaps) not intuitive, and therefore harder
to annotate by coders. It would be useful to investigate whether these
labels can be inferred, in such a way that constraints with colons
(for negation and conditionals) could be simplified.  In what
direction could this go? First note that inferred contexts are needed
in cases of negation (and conditionals). In DRT, a negation is formed
recursively, where the negation material (represented as a DRS) is
embedded into the wider context (again, represented as a DRS). But
when there is nothing available in the wider context, the
corresponding DRS will be empty (and as a result, there won't be a
corresponding labelled AMR). As we proposed in
Section~\ref{sec:negation}, we infer an index to meet the requirements
of a well-formed logical structure.
Instead, we can adopt a short-hand notation for such cases, i.e.,
$\{\lnot j\}$ meaning $\{i:\lnot j\}$, and $\{ j \Rightarrow k\}$
meaning $\{ i: j \Rightarrow k\}$.
But in general, conditionals would require an equivalent notation in
terms of negation, so $\{ \lnot j, j: \lnot k\}$ as short for $\{i:
\lnot j, j: \lnot k\}$, which is logically equivalent to $\{ i: j \Rightarrow k\}$.
This would cover all cases in Section~\ref{sec:quant}.

\subsection{Annotation Work}

Existing AMR annotations can be monotonically extended: all
``slashes'' that occur in AMRs need to be indexed, and constraints
need to be added.  Given an annotated AMR corpus \cite{amr}, this can
be done semi-automatically: first add indices automatically (by
replacing "/" by "/1/"). Then manually correct cases of negation
(search for ":polarity -"), universal quantification, and definite
descriptions, for a sample of the corpus. Finally, use machine
learning to annotate the rest (or hire or persuade human annotators to
do the job).
The polarity and mod relations can be optionally removed from the AMRs.
An interesting question concerns the use of existing parsing models
developed for AMR \cite{artzi-etal-2015-broad,NoordBos2017,damonte-etal-2017-incremental}. How can
these be extended to deal with the extended AMRs as proposed in this
paper? As the original AMR is not affected, a sensible approach would
be one in a modular fashion, where datasets consisting of AMRs paired
with AMR$^+$s could be used as training material.

\subsection{Triple Format of Logical Structure}

AMRs are converted to sets of triples for evaluation
purposes. Therefore, a sensible question to ask is how the logical
constraints in this proposal are converted to triples. There are two new types of triple instances.
 The indexed
sub-AMRs all introduce a membership triple (with the edge named IN) linking an instance with a scope
index.  Secondly, each scoping constraint introduces one or more
``structural'' triples. Constraints that involve two indices introduce a single
triple. Constraints that involve three indices introduce two triples. Here is an example:

\begin{quote}
"Nobody smiled."\\
\amr[ $\{$4:2$=>$3,3:$\lnot$1$\}$]{e}{1}{smile-01 \rol{26}{ARG0}{\amr{x}{2}{person}}}
\end{quote}

This will introduce the membership triples 
$<$ e , IN , 1 $>$ and
$<$ x , IN , 2 $>$, the triple for negation
$<$ 3 , NOT , 1 $>$, and the triples for the implication
$<$ 4 , IF , 2 $>$, and
$<$ 2 , THEN , 3 $>$.
As a consequence, standard tools for AMR evaluation \cite{cai-knight-2013-smatch} can be used, or extended to more fine-grained scores \cite{damonte-etal-2017-incremental}.

\section{Conclusion and Future Work}

The original AMR notation can be extended by a layer of logical
structure that gives correct interpretation of linguistic phenomena
that require scope or quantification.  The resulting framework bears
strong similarities with Discourse Representation Theory as
implemented in the Parallel Meaning Bank, and it deals with the
\textit{bound variable problem} and the \textit{scope representation
  problem}. So what's next? Let's get an AMR corpus annotated with
logical structure!

\section*{Acknowledgements}

This work was funded by the NWO-VICI grant \textit{Lost in Translation
  -- Found in Meaning} (288-89-003).  I thank the three anonymous
reviewers for their feedback.  Reviewer~1 rightly reminded me to cite
\newcite{artzi-etal-2015-broad}, and so I did; and I also followed
their suggestion to say more about the inferred contexts
(they can now be found in Section~\ref{sec:inferred}), and include more negation examples in
Section~\ref{sec:negation} (I added "no woman smiled").  Reviewer~2 was under the incorrect assumption that I was
unaware of \newcite{Crouch:2018}, but actually, I already referred to this article
in the submitted version of this paper, which indeed shares many
points of contact with my work.  Reviewer~3 had some excellent suggestions, and
also requested a formal definition of the translation, and I thought this was
a great idea. I somehow managed to squeeze it in this short paper, forcing
me to leave out some of the details.  Thanks again!

\bibliographystyle{coling}

\end{document}